\documentclass[lettersize,journal]{IEEEtran}
\usepackage{amsmath,amsfonts}
\usepackage{array}
\usepackage[caption=false,font=normalsize,labelfont=sf,textfont=sf]{subfig}
\usepackage{textcomp}
\usepackage{stfloats}
\usepackage{pifont}
\usepackage{url}
\usepackage{verbatim}
\usepackage{graphicx}
\usepackage{amssymb}
\usepackage{booktabs}
\usepackage{multirow}
\usepackage{comment}
\usepackage{xcolor}
\newtheorem{remark}{Remark}

\usepackage{soul}
\usepackage{hyperref}
\usepackage[framemethod=tikz]{mdframed}
\usepackage{enumitem}
\usepackage{amssymb}
\usepackage{algpseudocode}
\usepackage{algorithm}
\usepackage{makecell}

\makeatletter
\def\endthebibliography{%
	\def\@noitemerr{\@latex@warning{Empty `thebibliography' environment}}%
	\endlist
}
\makeatother

\begin{document}
\newcommand{\EE}{E2E}

\title{DSDrive: Distilling Large Language Model for Lightweight End-to-End Autonomous Driving with Unified Reasoning and Planning
}

\author{
    Wenru Liu,   
    Pei Liu,
    	and Jun Ma, \textit{Senior Member, IEEE}
    \thanks{Wenru Liu, Pei Liu, and Jun Ma are with The Hong Kong University of Science and Technology, China (e-mail: wliu354@connect.hkust-gz.edu.cn; pliu061@connect.hkust-gz.edu.cn;  jun.ma@ust.hk).} 
	}



\maketitle

\begin{abstract}  
In conventional end-to-end frameworks for Autonomous driving (AD), the underlying cognitive processes are inadequately addressed. While large language models (LLMs) offer improved understanding and reasoning capabilities, integrating them into AD systems poses two major challenges. First, there is an invariably significant discrepancy between the high computational demands of LLMs and the high efficiency required for autonomous vehicles. Second, although LLMs can generate high-quality semantic reasoning, it remains an open challenge to map high-level textual reasoning to low-level trajectory planning for autonomous vehicles. 
To deal with these issues, we present DSDrive, a streamlined end-to-end paradigm tailored for integrating the reasoning and planning of autonomous vehicles into a unified framework. DSDrive leverages a compact LLM that employs a distillation method to preserve the enhanced reasoning capabilities of a larger-sized vision language model (VLM). 
To effectively align the reasoning and planning tasks, a waypoint-driven dual-head coordination module is further developed, which synchronizes dataset structures, optimization objectives, and the learning process.
By integrating these tasks into a unified framework, DSDrive anchors on the planning results while incorporating detailed reasoning insights, thereby enhancing the interpretability and reliability of the end-to-end pipeline.
DSDrive has been thoroughly tested in closed-loop simulations, where it performs on par with benchmark models and even outperforms in many key metrics,  all while being more compact in size. Additionally, the computational efficiency of DSDrive (as reflected in its time and memory requirements during inference) has been significantly enhanced. Evidently thus, this work brings promising aspects and underscores the potential of lightweight systems in delivering interpretable and efficient solutions for AD.
\end{abstract} 


Video of the experiments: ~\href{https://youtu.be/op8PzQurugY}{https://youtu.be/op8PzQurugY}



\section{Introduction}
    Autonomous vehicles (AVs) have demonstrated pervasive power in modern intelligent transportation systems, which significantly enhances on-road driving safety and travel efficiency~\cite{mozaffari2020deep}. 
    The end-to-end ({\EE}) autonomous driving (AD) framework~\cite{chen2022milestones}, leveraging large-scale driving datasets~\cite{guo2019safe} and advanced learning methodologies~\cite{muhammad2020deep}, plays an important role in advancing the forefront of AVs.
    This fully differentiable framework processes raw sensor inputs to output low-level control through joint task optimization.
    In recent years, large language models (LLMs)~\cite{anthropic2024claude,achiam2023gpt} present new frontiers for the development of AVs~\cite{liu2025codrivevlm}. In this context, the application of LLMs to AD, particularly concerning their integration with the \EE\ AD framework, has garnered substantial research interest~\cite{liu2025vlm}.
    Essentially, LLMs effectively address the limitations of the traditional \EE\ AD framework, which lacks a genuine understanding of the driving task and is deficient in explainability~\cite{xiao2020multimodal}. 
    This is because LLMs provide a compelling solution through their cognitive processing capabilities and language generation abilities. The former enables nuanced contextual reasoning for the driving task~\cite{nie2024reason2drive}, while the latter facilitates a natural human-computer interaction interface.
    However, deploying an LLM-based AD framework remains challenging.
    Intrinsically, the computational demands of large-scale language models create substantial bottlenecks in real-time decision-making, and this imposes substantial constraints on the memory and processing latency of critical planning functions~\cite{gong2023edge}.
    
    While downsizing the models provides an underpinning basis, the current research efforts demonstrate  that the capabilities of LLMs are generally related to model size~\cite{zhou2024vision}, which implies that directly employing compact-sized architectures may compromise essential cognitive functions required for safe autonomous operations.
    Recent advances in model optimization offer promising pathways forward. Notably, knowledge distillation techniques exemplified by DeepSeek-R1~\cite{guo2025deepseek} demonstrate that sophisticated reasoning capabilities can be effectively transferred from large-sized teacher models to compact-sized student architectures. This approach enables the creation of compact yet capable LLM implementations that maintain sufficient decision-making competence while addressing computational constraints.
    This implication holds significant relevance for the application of LLMs in the context of of E2E AD. The core challenge is to design efficient knowledge transfer mechanisms that can endow compact models with advanced cognitive capabilities beyond their inherent performance limitations; and this can be achieved through strategic architectural design and targeted training paradigms.
    Essentially, knowledge distillation has proven effective in reconciling discrepancies in model size and capability. 
    However, it does not resolve the inherent discrepancy between the high-level textual reasoning offered by LLMs and the low-level trajectory planning required by AVs~\cite{ma2023eureka}. 
       While LLMs excel at semantic reasoning, they are not optimized for the numerical computation and spatial reasoning tasks that are essential for trajectory planning in AD{~\cite{chen2024driving}}.    
    This mismatch has led most existing LLM-based AD research to rely on open-loop evaluations~\cite{huang2024drivemm}. The transition of LLM-based \EE\ AD frameworks to closed-loop planning is imperative for practical AD deployment~\cite{zhang2022rethinking}. The \EE\ AD framework serves an effective approach to coordinate different tasks toward closed-loop planning. By focusing on the ultimate driving objective, the planning result aligns various tasks within a unified framework, which ensures that they can collectively advance toward the shared goal~\cite{liu2024integrating}. Motivated by this, it leaves an interesting problem to integrate LLMs into the \EE\ AD framework from a planning-oriented perspective~\cite{lu2024activead}.
    
    In this work, we propose a novel lightweight \EE\ AD framework to tackle the challenges of computational efficiency and the disconnection between reasoning and planning. This framework leverages compact LLMs to achieve high reasoning and planning performance comparable to large-sized systems. Our approach hinges on two key strategies: Firstly, we utilize a distillation method to enhance the compact LLM's capabilities as the core of our AD system. This entails externalizing the reasoning power of a large-sized Vision Language Model (VLM) via a chain-of-thought (CoT) prompting process~\cite{yao2024calmm}. We generate structured datasets that explicitly capture think-and-answer reasoning, enabling the transfer of reasoning skills from foundational models to specialized systems. 
    We propose a waypoint-driven dual-head coordination module that facilitates the appropriate alignment of reasoning and planning tasks. 
        This module entails the co-design of the training dataset, ensuring that both tasks are grounded in homogeneous data. It innovatively incorporates the planning outcome as the final answer for the reasoning process, thereby establishing a unified optimization objective for the two tasks. The two tasks are trained jointly within the E2E framework, simultaneously optimizing for the quality of the reasoning answers and the accuracy of waypoint predictions. By interlinking these tasks, our framework supports closed-loop AD with explicit reasoning, thereby enhancing the explainability and reliability of autonomous systems.
    The contributions of our work are summarized as follows:
    \begin{itemize}
        \item We present DSDrive, a lightweight E2E AD framework that employs a compact LLM to process multi-modal inputs for explicit reasoning and closed-loop planning. Specifically, we utilize knowledge distillation to empower the compact LLM to undertake the reasoning and planning tasks, thereby improving its overall performance.
   
        \item We present a novel waypoint-driven dual-head coordination module that bridges high-level reasoning and low-level trajectory planning. By integrating waypoints into the explicit reasoning process, we establish a unified objective for both tasks and facilitate their collaborative progression toward the overarching driving goal.

        
        \item We perform extensive closed-loop simulations in CARLA to showcase the efficacy of knowledge distillation and the dual-head coordination module. The proposed DSDrive achieves driving performance comparable to larger-sized benchmark models, and even outperforms them in several key metrics, particularly in terms of enhanced computational efficiency.

    \end{itemize}
    
    The remainder of this paper is organized as follows: Section \ref{sec:related} delves into the related work, offering essential background on \EE\ AD and the application of LLMs in AD. Section \ref{sec:method} outlines the methodology, detailing the framework components and the distillation-based training scheme. In Section \ref{sec:exp}, we present the experiments and pertinent results, which vividly demonstrate the effectiveness and performance of our proposed approach. Finally, Section \ref{sec:con} provides concluding remarks and discusses potential directions for future research.

\section{Related Work} \label{sec:related}
    \subsection{End-to-End Autonomous Driving}
    One of the motivations behind the {\EE} AD framework is to develop a unified approach that integrates perception, decision-making, and planning tasks, optimizing the entire pipeline for the ultimate driving performance. This goal has been pursued through innovative architectural approaches. ST-P3 \cite{hu2022st} advances spatial-temporal feature learning to unify scene understanding across multiple driving tasks. Building on this foundation, UniAD \cite{hu2023planning} establishes an integrated framework that synergizes perception and prediction modules to enhance planning capability. 
    Architectural innovations continue to reshape system design, aiming to advance the {\EE} AD framework with more resourceful skills. DriveTransformer \cite{jia2025drivetransformer} achieves unified task interaction through parallel processing architectures with sparse representations, improving operational stability. Inspired by cognitive mechanisms, the cascaded decision-making framework incorporates multi-stage safety verification for action refinement \cite{jia2023think}. Complementary approaches include a planning-centric coordination system \cite{Hu2022goal} and a vision-based multi-agent collaboration model \cite{cui2022coopernaut}, which enhances emergency response through cross-vehicle perception. ReasonNet \cite{shao2023reasonnet} implements hierarchical reasoning architectures for improved scenario understanding and traffic participant behavior prediction.
    The research community has concurrently explored alternative paradigms to pushing the limit of the {\EE} AD framework. VAD \cite{jiang2023vad} investigates vectorized scene representations to reduce map dependency in motion planning, while its successor VADv2 \cite{chen2024vadv2} introduces probabilistic planning strategies and environmental tokenization for enhanced uncertainty handling. 
    
    Despite these innovations, a fundamental limitation of the {\EE} approach is its lack of structured reasoning mechanisms to decompose complex scenarios into logical decision pathways. This shortcoming can be ascribed to two perspectives. Firstly, the traditional deep learning network does not have intrinsic capabilities to perform sophisticated reasoning or interpretation. Secondly, the reasoning patterns necessary for the system to learn a comprehensive understanding are typically absent in the training data. As discussed in many previous works~\cite{bansal2018chauffeurnet, cheng2024pluto, lu2023imitation}, the acquisition of reasoning ability cannot rely on merely scaling up the size of neural network models and data. Therefore, it necessitates a paradigm shift towards integrating structured reasoning mechanisms in both model architecture and training data to address the limitations of the current \EE\ pipeline for AD.

    \subsection{LLMs in Autonomous Driving}
    It is readily reported in the recent works that the integration of LLMs in the \EE\ AD framework has demonstrated advancements in environmental reasoning and decision-making. To address interpretability challenges pertinent to \EE\ AD, RAG-Driver \cite{yuan2024rag} employs retrieval-augmented multimodal LLMs to generate human-understandable explanations for driving actions while maintaining precise control signal prediction, demonstrating notable zero-shot generalization across unfamiliar environments.
    In planning optimization, AlphaDrive \cite{jiang2025alphadrive} combines GRPO-based reinforcement learning with reasoning strategies to enhance training efficiency and multimodal planning capability. PRIMEDrive-CoT \cite{mandalika2025primedrive} advances safety-critical decision-making through Bayesian graph neural networks integrated with CoT reasoning, employing visual attention mechanisms to improve transparency in risk assessment. GPT-Driver~\cite{mao2023gpt} explores human-AI interaction paradigms using ChatGPT-3.5 for natural language command interpretation in trajectory planning scenarios.

    Recent advancements in integrating LLMs into AD systems have yielded several notable frameworks. DriveLM~\cite{sima2024drivelm} pioneers a multimodal fusion architecture that aligns linguistic reasoning with sensor inputs through CoT prompting. VLP~\cite{pan2024vlp} employs vision-language pre-training to enhance scene understanding. WiseAD~\cite{zhang2024wisead} introduces knowledge distillation to compress LLM knowledge into lightweight policy networks. Sce2DriveX~\cite{zhao2025sce2drivex} proposes a scenario-aware encoder-decoder framework that improves generalization across driving conditions. LMDrive~\cite{shao2024lmdrive} develops an {\EE} architecture with spatiotemporal tokenization for trajectory prediction, but it lacks an explicit reasoning mechanism within its design. 
    As summarized by Table~\ref{tab:llm_ol}, although these approaches have shown progress in open-loop evaluations by improving multimodality alignment in perception and reasoning, they collectively fail to adequately address closed-loop driving scenarios. The closed-loop driving demands alignment between reasoning and planning tasks, and also places stricter requirements on computational resources. This limitation underscores the critical need for developing an \EE\ AD system that harmonizes lightweight LLM deployment with robust closed-loop planning capabilities.

    \begin{table}[t]
      \centering
      \caption{Comparison of representative works in LLM-based AD.}
        \begin{tabular}{ccccc}
        \toprule
        Method & E2E   & Closed-loop &  Explicit reasoning \\
        \midrule
        DriveLM~\cite{sima2024drivelm} & \ding{55} & \ding{55} &  \ding{51} \\
        VLP~\cite{pan2024vlp}   &  \ding{55}     &   \ding{55}     &  \ding{55} \\
        WiseAD~\cite{zhang2024wisead} & \ding{51} & \ding{55}  & \ding{51} \\
        Sce2DriveX~\cite{zhao2025sce2drivex} & \ding{51} &   \ding{55}     & \ding{51} \\
        Senna~\cite{jiang2024senna} & \ding{51} &   \ding{55}     & \ding{51} \\
        LMDrive~\cite{shao2024lmdrive} & \ding{51} & \ding{51}       & \ding{55} \\
        \textbf{Ours}  & \ding{51} & \ding{51}        & \ding{51} \\
        \bottomrule
        \end{tabular}%
      \label{tab:llm_ol}%
    \end{table}%

\section{Methodology} \label{sec:method}

    As an overview, our method consists of two main components: a reasoning model detailed in Section~\ref{sec:right}, and an \EE\ driving model outlined in Section~\ref{sec:left}. The reasoning model leverages a large-sized VLM, while the driving model utilizes a compact-sized LLM. Our method employs distillation as the learning mechanism, enabling the compact-sized driving model to acquire reasoning capabilities from the VLM, as elaborated in Section~\ref{sec:distill}.
 
    \subsection{Reasoning Model} \label{sec:right} 


    Our method utilizes two types of inputs: image $I$ and text $X$. The image input is a sequence represented as $I \in \mathbb{R}^{T \times H \times W \times C}$, where $T$ is the sequence length, $H$, $W$, and $C$ represent the height, width, and number of channels, respectively. The textual inputs are subdivided into navigation instructions and questions.

    At a high level, the process of utilizing the VLM in the reasoning model can be defined as:
    \begin{align} \label{eq:vlm}
       A_{vlm} = V (I, X_{navi}, X_{prompt}), 
    \end{align}
    where $V(\cdot)$ represents the VLM, which processes visual and textual inputs to generate answers $A_{vlm}$. 
    To enable the distillation of reasoning capabilities from the VLM to the driving model, we ensure the VLM processes driving image sequences and produces multi-faceted analyses aligned with the driving model's input structure.

    Specifically, we employ Qwen2.5-VL-max~\cite{bai2025qwen2}, a state-of-the-art open-sourced VLM, to generate a structured reasoning dataset regarding scene understanding, key object identification, and informed driving decision-making. As Qwen2.5-VL-max is a general-purpose VLM designed to handle a wide range of tasks, we implement a structured CoT strategy to refine its specialization for AD, with the following considerations:
    \begin{enumerate}
        \item \textbf{Scenario Comprehension:} Grasp the driving environment by analyzing elements such as weather, time of day, road type, and road conditions.
        \item \textbf{Key Item Description:} Detail the key item by its name, position, features, and the potential risks it poses to the ego vehicle's operation.
        \item \textbf{Strategic Driving Plan Formulation:} Develop a well-thought-out driving plan to navigate the scenario effectively.
        \item \textbf{Human-Interpretable Explanation Generation:} Create understandable explanations for the driving decisions and actions taken.
    \end{enumerate}

        \begin{figure}
        \centering
        \includegraphics[trim=0 0 0 0, clip, width=1\linewidth]{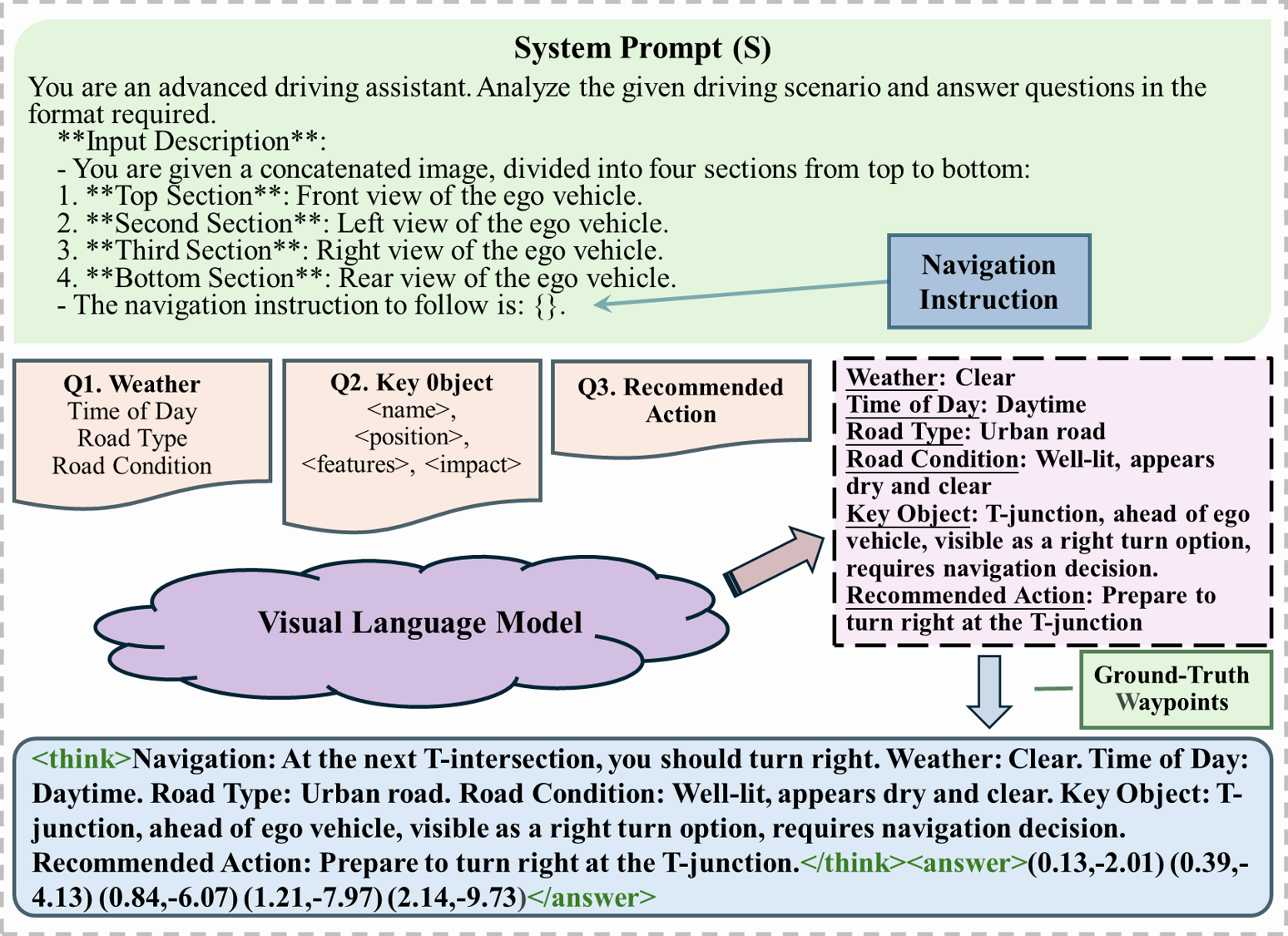}
        \caption{The dataset for distillation is prepared with an explicit think-and-answer process. This design enables the model to learn complex reasoning steps before generating a final answer, enhancing its learning of nuanced driving tasks. The dataset is diverse and realistic, covering various driving scenarios to improve model robustness and generalization.}
        \label{fig:prompt}
    \end{figure}

    Thus far, VLM has primarily provided insights centered on high-level reasoning. However, planning tasks in AD generally involve predicting waypoints for trajectory planning and subsequent vehicle control. Accurately forecasting target waypoints for precise maneuvers poses a significant challenge for general-purpose VLMs. This is mainly because spatial positioning has long been a recognized weakness in these models. Consequently, additional strategies are needed to bridge the gap between abstract reasoning and concrete planning tasks in AD.
    We have innovatively utilized ground-truth waypoints to connect the reasoning process to the planning task. Our inspiration comes from the training template in~\cite{guo2025deepseek}, which emphasizes a step-by-step thinking process followed by a final answer. This dataset design aims to prevent shortcuts to the answer by requiring the explicit expression of reasoning steps. In our work, this approach serves a more critical purpose. By embedding ground-truth waypoints within the think-and-answer process, we have seamlessly integrated the planning task as a natural outcome of the reasoning process.
    
    We summarize our approach in leveraging the reasoning model to prepare the training dataset in Fig.~\ref{fig:prompt}. The complete prompts for the VLM are defined as \( X_{prompt} = [S, Q_1, Q_2, Q_3] \), where \( S \) represents the system prompt, which outlines the task's user-assistant interactive setting along with the input-output structure. This is followed by a sequence of task-specific prompts, \( Q_1, Q_2, Q_3 \), which guide the VLM through coherent progression from perception to prediction and decision-making. The generated answer by the VLM is then concatenated with the navigation instruction and ground-truth waypoints to form a structured think-and-answer reasoning process, which constitutes the training dataset.
    Remarkably, the construction of our training dataset emphasizes the externalization of reasoning mechanisms that are inherent to the VLM but are typically absent in conventional driving models. Furthermore, it focuses on establishing a direct linkage between the reasoning process and the planning outcome.

\subsection{End-to-End Driving Model} \label{sec:left}
    \begin{figure*}[t]
        \centering
        \includegraphics[trim=0 0 0 0, clip, width=1\linewidth]{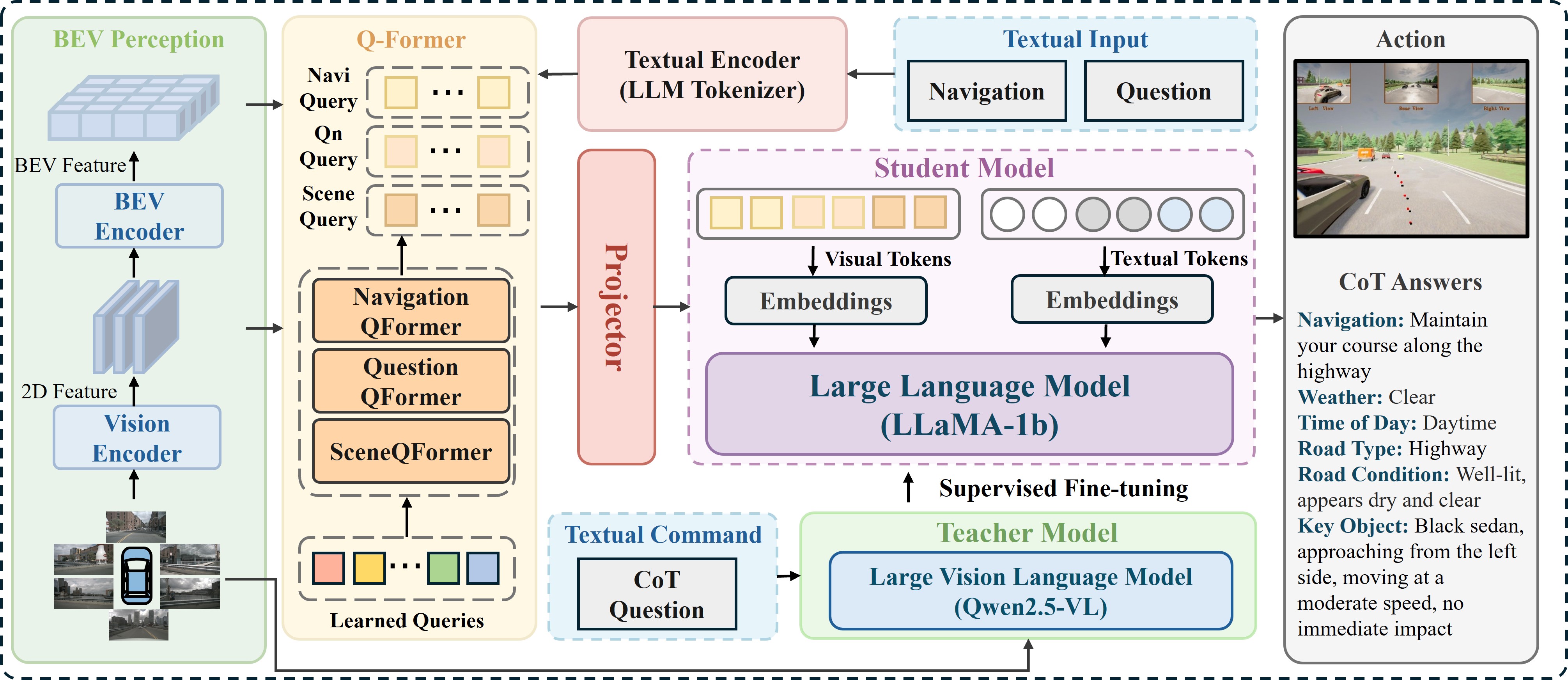}
        \caption{The proposed \EE\ driving model boasts advanced reasoning and planning capabilities. Its student model ensures inference efficiency and is enhanced via distillation from a teacher model, which substantially improves reasoning skills. By integrating reasoning and planning, it achieves robust performance and interpretability, making it highly reliable and accurate for AD tasks.}
        \label{fig:model}
    \end{figure*}

        The driving model takes an {\EE} approach, defined as follows:
    \begin{align} \label{eq:e2e}
       W_{pred},  A_{pred}, C_{pred} = E (I, X_{navi}, X_{qn}),
    \end{align}
    The driving model processes multimodal inputs, including visual data and textual information, to output predicted waypoints ($W_{pred}$), predicted answers ($A_{pred}$), and an indicator of whether the current navigation command has completed ($C_{pred}$).   
    The image inputs and navigation instructions are homogeneous to the reasoning model, ensuring consistency across the framework. 
    While the questions for the reasoning model, $X_{vlm}$, consist of a series of questions regarding subtasks in AD, the question text $X_{qn}$ in the driving model is a straightforward commanding sentence like \textit{``think and answer"}.

    As illustrated in Fig.~\ref{fig:model}, the driving model integrates several specialized components to enable comprehensive reasoning and planning: (1) Visual encoder and tokenizer for processing the inputs, (2) Q-former for aligning visual features and textual tokens, (3) LLM backbone, and (4) Multi-tasks outputs with a dual-head design of the reasoning and planning tasks.
    
    \subsubsection{Inputs Encoding}
        Both the image and text inputs are processed to prepare tokens for use by the LLM.
        We adopt a pre-trained vision encoder from~\cite{shao2024lmdrive}, which effectively transforms multi-view, multi-modality sensor data into embedded features. The image sequence \( I \in \mathbb{R}^{T \times H \times W \times C} \) is processed through the visual encoder to extract visual features \( F_v \in \mathbb{R}^{T \times N_v \times D_v} \), where \( N_v \) is the number of visual tokens and \( D_v \) is the hidden dimension. 
        Alongside the visual encoder, we apply the LLaMA tokenizer~\cite{grattafiori2024llama} to convert the textual inputs, including the navigation instruction and questions, into textual tokens \( H_l \in \mathbb{R}^{N_l \times D_l} \), where \( N_l \) denotes the length of the textual sequence and \( D_l \) is the hidden dimension of the language model embeddings. To ensure temporal alignment with the image sequence, text tokens are repeated \( T \) times, corresponding to each frame in the sequence.
    
    \subsubsection{Aligning Visual and Textual Modalities}
    A critical aspect of processing visual and textual inputs is ensuring effective cross-modal alignment. To achieve this, we employ the Q-Former architecture from BLIP-2~\cite{li2023blip}. The Q-Former strategically queries encoder outputs to align visual features with textual tokens, bridging the visual and text modalities. To enhance efficiency, we generate a set of learnable queries \( Q \in \mathbb{R}^{K \times D_l} \), where \( K \) is the number of query tokens. These tokens reduce the dimensionality of visual tokens, helping the model focus on relevant visual input aspects. Next, we apply a cross-attention mechanism with attention masks on query tokens, text tokens, and image embeddings. The Q-Former's output, specifically the last hidden states, passes through a linear layer. This projects the visual features into the language model's embedding space, ensuring compatibility with the LLM backbone.

    \subsubsection{LLM Backbone}   
        We employ LLaMA-1B as the LLM backbone, as compared to LLaMA-7B utilized in LMDrive. 
        The LLM backbone processes the integrated embeddings derived from images, navigation instructions, and questions. 
        To facilitate the alignment of reasoning and planning tasks through a multi-task design, we opt to decode the outputs using the hidden states \( H \), which contain contextual information from multiple modalities, rather than using the direct outputs from the LLM.

    \subsubsection{Multi-Task Outputs}
    Based on the hidden states \( H \) from the LLM, the driving model is designed with a multi-task output structure, enabling the simultaneous execution of several functions. It generates planned trajectories that can be transformed into precise vehicle maneuvers via waypoints prediction, provides natural language explanations of the step-by-step reasoning behind driving decisions, and determines whether the driving task under the current navigation is completed.
    
    \textbf{Waypoint Prediction}
    The waypoint predictor \( f_{wp} \) projects the hidden states to predict future vehicle positions:
    \[ W = f_{wp}(H), \]
    where \( f_{wp} \) is a simple feed-forward neural network with two MLPs and a ReLU activation in between.

    \textbf{CoT Reasoning Prediction}
    The CoT reasoning predictor, denoted as $f_{cot}$, is implemented using customized Qwen2-style decoder layers. The reasoning predictor is connected with the waypoint predictor such that both components share the same objective of planning for waypoints, and they both require supervision from ground-truth waypoints.
    As outlined in Algorithm~\ref{alg:cot_training} and Algorithm~\ref{alg:cot_inference}, we employ two strategies for reasoning generation during training and inference.
    In training, we leverage the teacher forcing strategy that is discussed in detail later in Section~\ref{sec:distill}. 
    When deployed in closed-loop simulations, the model uses an auto-regressive approach to generate answers for reasoning. This involves generating one token at a time, where each token is predicted based on the context provided by the previously generated tokens and the initial hidden states from the LLM.
 
    \begin{remark}
    As reasoning and planning are two closely correlated tasks in AD, we strategically develop a waypoint-driven dual-head coordination module for the alignment of the two tasks. This module embeds ground-truth waypoints as the final output of the reasoning process, ensuring that both the reasoning and driving predictors share a common objective. In this sense, the reasoning task also acts as a translator for the planning task by explicitly referencing the predicted waypoints, thereby enhancing the explainability of the E2E AD framework. 
    \end{remark}

    \textbf{End State Prediction}
    Similar to the waypoint predictor, the end state predictor \( f_{end} \) uses two MLPs with a ReLU activation in between to predict the probability of the ending position:
    \[ C_{pred} = f_{end}(H). \]

\subsection{Reasoning Distillation} \label{sec:distill}
    Within our proposed framework, the large-sized VLM is characterized as the reasoning model, as it comes with advanced reasoning capability and is expected to enhance the AD system by providing an explicit reasoning process. On the other hand, the driving model is responsible for the actual vehicle planning but possesses limited reasoning ability due to its compact size. A critical aspect of our framework is the distillation of reasoning capability from the reasoning model (i.e., a large-sized VLM) to the driving model (i.e., a compact-sized LLM). 
    
    To achieve this, a teacher-forcing approach (Algorithm~\ref{alg:cot_training}) is employed. 
    The distillation exposes the compact-sized driving model to the same input data as the large-sized reasoning model. The driving model is guided to produce reasoning answers that closely match those of the reasoning model.
    Based on a sequence of hidden states by the LLM backbone, the hidden states from the last few key frames are extracted and concatenated with the embedded tokens of the ground-truth answers to facilitate the teacher-forcing training. This combined sequence is then passed through the answer predictor, which generates predictions for each token in the sequence.
    This training process aims to minimize the divergence between the answers predicted by the driving model and those provided in the training template, which consists of answers generated by the reasoning model alongside ground-truth data. The objective is to enable the driving model to internalize the reasoning strategies demonstrated by the large-sized model, thereby enhancing its reasoning capability.
    
    Only the driving model is involved in the training process. Cross-entropy loss between the predicted logits and the target tokens is computed and used to optimize the answer predictor. The teacher-forcing approach ensures that the model learns to generate coherent and precise explanations by directly conditioning on the ground-truth sequence during training. 
    
    In total, three loss terms are defined based on the outputs from the driving model.
    \begin{itemize}
        \item $loss_{wp}$: The waypoint loss is an $l_1$ loss of the predicted waypoint and the ground-truth waypoint.
        \item $loss_{ans}$: The answer loss is computed using cross-entropy loss between predicted and ground-truth answers.
        \item $loss_{end}$: This loss checks whether the input instruction is completed, and we employ cross-entropy loss for this classification task.
    \end{itemize}
    
    Therefore, the training objectives are defined through the following loss function:
    \begin{align} \label{eq:loss}
       loss = \lambda_1 loss_{wp} + \lambda_2 loss_{ans} + \lambda_3 loss_{end},
    \end{align}
    where $\lambda_1$, $\lambda_2$, and $\lambda_3$ are weighting coefficients in the range of zero to one that control the importance of each component to the joint loss.

\begin{algorithm}[t]
\caption{Reasoning Training with Teacher Forcing}\label{alg:cot_training}
\begin{algorithmic}[1] 
\State \textbf{Input:} Output context $H$ from the LLM, ground truth answer $A$, maximum sequence length $L$
\State \textbf{Return:} Training loss $\mathcal{L}$

\State Tokenize $A$ to get $\text{cot\_answer\_token}$
\If{$\text{cot\_answer\_token}[0, -1] \neq \text{EOS}$}
\State Append EOS token to $\text{cot\_answer\_token}$
\EndIf
\State $\text{target\_labels} \gets \text{cot\_answer\_token}[:, 1:]$ 
\State $\text{cot\_answer\_token} \gets \text{cot\_answer\_token}[:, :-1]$ 

\State Embed $\text{cot\_answer\_token}$ to get $\text{cot\_answer\_embeds}$
\State Concatenate $H$ and $\text{cot\_answer\_embeds}$ to form $\text{sequence}$
\State Create attention mask $M$ and position IDs $P$ for $\text{sequence}$
\State $\text{outputs} = f_{\text{cot\_predictor}}(\text{sequence}, M, P)$ \Comment{Answer prediction using Qwen2\-style decoder layers}
\State Extract hidden states: $\text{hidden\_states} = \text{outputs}[0]$

\State $\text{context\_length} \gets \text{len}(H)$
\State $\text{dec\_output} \gets \text{hidden\_states}[:, \text{context\_length}:, :]$ 
\If{$\text{len}(\text{dec\_output}) \neq \text{len}(\text{target\_labels})$}
\State $\text{min\_len} \gets \min(\text{len}(\text{dec\_output}), \text{len}(\text{target\_labels}))$
\State $\text{dec\_output} \gets \text{dec\_output}[:, :\text{min\_len}, :]$ 
\State $\text{target\_labels} \gets \text{target\_labels}[:, :\text{min\_len}]$
\EndIf

\State $\text{predicted\_logits} \gets f_{\text{proj}}(\text{dec\_output})$ \Comment{Project output to logits}
\State $\mathcal{L} = \text{CrossEntropy}(\text{predicted\_logits}, \text{target\_labels})$

\State \Return $\mathcal{L}$
\end{algorithmic}
\end{algorithm}

\begin{algorithm}
\caption{Reasoning Inference with Autoregression}\label{alg:cot_inference}
\begin{algorithmic}[1]
\State \textbf{Input:} Output context $H$ from the LLM, maximum number of tokens $N$
\State \textbf{Return:} Generated answer $\hat{A}$
\State \textbf{Initialization}
\State \indent $\text{generated\_tokens} \gets []$
\State \indent $\text{current\_token} \gets \text{BOS}$
\State \indent Embed $\text{current\_token}$ to get $\text{current\_embed}$

\For{$\text{step} = 1$ \textbf{to} $N$} \Comment{Autoregression}
    \State Combine $H$ and $\text{current\_embed}$ as $\text{full\_sequence}$
    \State Create attention mask $M$ and position IDs $P$
    \State $\text{outputs} = f_{\text{cot\_predictor}}(\text{full\_sequence}, M, P)$
    \State $\text{last\_hidden\_state} = \text{outputs}[:, -1, :]$
    \State $\text{logits} = f_{\text{proj}}(\text{last\_hidden\_state})$
    \State Apply repeat penalty to $\text{logits}$
    \State Greedy sampling: $\text{next\_token} = \arg\max(\text{logits})$
    \State Append $\text{next\_token}$ to $\text{generated\_tokens}$
    \If{$\text{next\_token} = \text{EOS}$}
        \State \textbf{break}
    \EndIf
\EndFor

\State Decode $\text{generated\_tokens}$ into a readable answer $\hat{A}$

\State \Return $\hat{A}$ 
\end{algorithmic}
\end{algorithm}

\section{Experiment and Result} \label{sec:exp}
    We hypothesize that by focusing on key reasoning patterns, the \EE\ driving model using compact LLM can acquire essential reasoning capability through distillation, thereby enhancing overall closed-loop driving performance while avoiding substantial computational overhead.
    Centered around this research motivation, the experiments are designed to provide a comprehensive evaluation of the proposed method along three key aspects: its quantitative and qualitative performance, the effectiveness of the waypoint-driven dual-head coordination module, and its computational efficiency.

\begin{table*}[htbp]
  \centering
  \caption{Closed-loop driving performance of DSDrive in CARLA simulations, with comparison to baseline and ablation methods.}
    \begin{tabular}{cccccccc}
    \toprule
          &  \multirow{2}{*}{Model}     & {\multirow{2}{*}{DS $\uparrow$}} & {\multirow{2}{*}{RC $\uparrow$}} & {\multirow{2}{*}{IS $\uparrow$}} & \multicolumn{3}{c}{Normalized number of critical incidents} \\
          \cline{6-8} 
          &       &       &       &       & \makecell[c]{Collision $\downarrow$} & \multirow{1}{*}{Red light violation $\downarrow$} & \multirow{1}{*}{Outside lanes $\downarrow$}\\
    \midrule
    \multirow{4}[1]{*}{LangAuto-Long} & Vanilla (LLaMA-1B) & 21.27  & 23.03  & 0.93  & 0.00  & 0.03  & 0.16  \\
          & LMDrive (LLaMA-1B) & 21.59  & 27.52  & 0.87  & 0.19  & 0.06  & 0.13  \\
          & LMDrive (LLaVA-7B) & 28.51  & 37.87  & 0.80  & 0.13  & 0.22  & 0.25  \\
          & DSDrive (LLaMA-1B) & 29.57  & 39.31  & 0.77  & 0.12  & 0.24  & 0.32  \\
          \midrule
    \multirow{4}[0]{*}{LangAuto-Short} & Vanilla (LLaMA-1B) & 38.75  & 42.88  & 0.87  & 0.07  & 0.14  & 0.21  \\
          & LMDrive (LLaMA-1B) & 41.86  & 48.71  & 0.88  & 0.07  & 0.14  & 0.21  \\
          & LMDrive (LLaVA-7B) & 45.03  & 52.90  & 0.84  & 0.07  & 0.20  & 1.00  \\
          & DSDrive (LLaMA-1B) & 62.05  & 76.18  & 0.81  & 0.07  & 0.27  & 1.00  \\
          \midrule
    \multirow{4}[1]{*}{LangAuto-Tiny} & Vanilla (LLaMA-1B) & 47.76  & 53.70  & 0.84  & 0.06  & 0.00  & 0.19  \\
          & LMDrive (LLaMA-1B) & 49.97  & 53.57  & 0.88  & 0.25  & 0.00  & 0.06  \\
          & LMDrive (LLaVA-7B) & 61.91  & 73.19  & 0.84  & 0.19  & 0.06  & 0.13  \\
          & DSDrive (LLaMA-1B) & 60.67  & 72.53  & 0.84  & 0.06  & 0.00  & 0.25  \\
    \bottomrule
    \end{tabular}%
  \label{tab:cl_metrics}%
\end{table*}%

    \subsection{Implementation Details}
    \subsubsection{Training}
    DSDrive is trained on the dataset generated in Section~\ref{sec:right}, containing $11067$ clips.
    The model is trained for $50$ epochs using a batch size of $1$. A linear warm-up followed by a cosine learning rate decay schedule is employed to facilitate stable and efficient optimization. During the initial $2000$ iteration steps, a warm-up phase is used to gradually increase the learning rate from $1\times10^{-6}$ to the base learning rate of $1\times10^{-4}$, thereby stabilizing the early stages of training. After the warm-up phase, the learning rate decays following a cosine schedule, reaching a minimum value of $1\times10^{-5}$ by the end of training. To mitigate overfitting, a weight decay of $0.06$ is applied. The training is conducted on 4 NVIDIA RTX 3090 with 24 GB of memory.    
    
    \subsubsection{Closed-Loop Simulations}
    The closed-loop evaluation is implemented using the CARLA simulator~\cite{dosovitskiy2017carla}. 
    The performance of DSDrive is compared with LMDrive~\cite{shao2024lmdrive}, the benchmark method dedicated to closed-loop \EE\ AD. 
    The efficacy of our proposed approach is also assessed utilizing the LangAuto benchmark provided by~\cite{shao2024lmdrive}. LangAuto is a collection of designated routes in CARLA that covers all eight publicly available towns with various kinds of environmental and weather conditions.
    In the CARLA simulator, the closed-loop evaluation operates at a frequency of 10 Hz. The proposed \EE\ AD model, DSDrive, processes sensor data, including images captured from the front, left, right, and rear cameras, as well as LiDAR point clouds. It then generates the reasoning process and predicted waypoints. Subsequently, these predicted waypoints are translated into control variables of steering, throttle, and brake inputs within the CARLA simulator.

    \subsubsection{Metrics}
    We consider the following metrics for the evaluation of driving performance in closed-loop: Route Completion (RC), Infraction Score (IS), and Driving Score (DS). Note that RC is the percentage of the route distance completed. A higher RC indicates a better completion of the driving task. IS quantifies infractions such as collisions, lane violations, traffic signal breaches, and route deviations. Starting at a base of $1.0$, the IS is reduced by each incident according to its severity. A higher IS suggests better adherence to traffic rules and safer driving practices. DS serves as a composite metric, capturing both driving progress and safety. It represents a balanced view of route navigation and regulations compliance. A higher DS is desirable. 
    Additionally, we count the number of critical incidents, such as collisions with vehicles, red light violations, and driving outside of lanes. These counted values are normalized to the number of simulated scenarios.

    \subsection{Benchmark Comparison}
    We compare the performance of the proposed method, DSDrive, in closed-loop simulations with the following alternatives:
    \begin{enumerate}
        \item LMDrive (LLaVA-7B): We directly employ the checkpoint provided by LMDrive~\cite{shao2024lmdrive}, which is the SOTA method in LLM-based closed-loop E2E AD.
        \item LMDrive (LLaMA-1B): This tiny version of LMDrive is based on a backbone LLM of LLaMA-1B, and is trained under the same strategy and dataset as LMDrive. It is designed to demonstrate the capability of a lightweight LLM like LLaMA-1B when integrated into \EE\ AD scenarios. It acts as an ablation study of DSDrive, validating that the integrated reasoning capability is effective in enhancing closed-loop driving performance.
        \item Vanilla (LLaMA-1B): This vanilla version applies LLaMA-1B directly within the \EE\ AD framework without any fine-tuning. It is intended to elucidate the intrinsic capability of the original compact-sized LLM in driving tasks.
    \end{enumerate}
    
    Our method, DSDrive, along with the comparative models, is evaluated using the LangAuto benchmark, which encompasses three distinct types of routes: LangAuto-Long, featuring routes longer than 500 meters; LangAuto-Short, comprising routes between 150 and 500 meters in length; and LangAuto-Tiny, consisting of routes shorter than 150 meters. 
    The testing results for DSDrive, along with the comparison methods, across the three LangAuto benchmarks, are presented in Table~\ref{tab:cl_metrics}.
    The comparative study is centered on one theme: a compact-sized LLM is inherently limited in its capabilities, and a conventional training strategy can only improve its performance to a certain extent, while our method is able to boost its performance to surpass that of a system with larger size.

    Starting with Vanilla (LLaMA-1B) and LMDrive (LLaMA-1B), which are based on the same LLM backbone as DSDrive. 
    Firstly, it is essential to recognize that LLaMA-1B, due to its compact size, inherently exhibits limitations when adapting to the complicated closed-loop driving task. The untrained model (Vanilla) records the lowest performance scores across all three benchmarks. The results are consistent with our hypothesis that the direct utilization of a compact-sized LLM imposes a substantial constraint on the overall performance of the system.    
    Secondly, using the same LLaMA-1B backbone, LMDrive applies its training strategy, but only marginally enhances the capabilities of LLaMA-1B. It fails to surpass the performance of large-sized systems, as demonstrated by LMDrive (LLaVA-7B).
    Finally, our method endows the compact-sized LLM with reasoning capability distilled from a large-sized VLM. This enhancement effectively complements the planning task and results in a significant improvement in the overall driving performance. This underscores the efficacy of integrating reasoning abilities into compact-sized LLMs as a means to augment AD performance.

    Specifically, the performance of DSDrive compared to the baseline method, LMDrive (LLaVA-7B), demonstrates the effectiveness of DSDrive in \EE\ AD, and highlights its reliability and consistency across varying route lengths. In the aspects of DS and RC, DSDrive closely trails LMDrive (LLaVA-7B) in the LangAuto-Tiny scenario. It shows marginal improvements over the baseline in the LangAuto-Long scenario and exhibits remarkable proficiency in the LangAuto-Short scenario. The IS of DSDrive is slightly lower than the baseline, but still maintains at a closely comparable level. The performance in IS could be further broken down into undesirable driving behaviors. DSDrive demonstrates competitive performance in minimizing collision rates and red light violations. However, it records higher lane deviations, which suggests a need for targeted enhancements to improve lateral stability and adherence to lane discipline.

    In summary, DSDrive demonstrates notable strengths in DS and RC across a variety of scenarios, showing its potential as an effective \EE\ AD agent. Its performance in collision avoidance and traffic rule compliance is commendable, although the higher frequency of lane deviations suggests an area for improvement. Importantly, the ablative studies comparing DSDrive with LMDrive (LLaMA-1B) and Vanilla (LLaMA-1B) substantiate the efficacy of our approach: by integrating advanced reasoning capability into compact-sized LLMs through distillation, we can enhance their performance in \EE\ closed-loop AD systems. Furthermore, the comparison with the baseline model, LMDrive (LLaVA-7B), highlights that, even with a compact-sized backbone model, our method has the potential to elevate the performance of compact-sized LLMs to the level of well-trained systems utilizing large-sized models.

\begin{figure*}[t]
    \centering
    \includegraphics[trim=0 0 0 0, clip, width=1\linewidth]{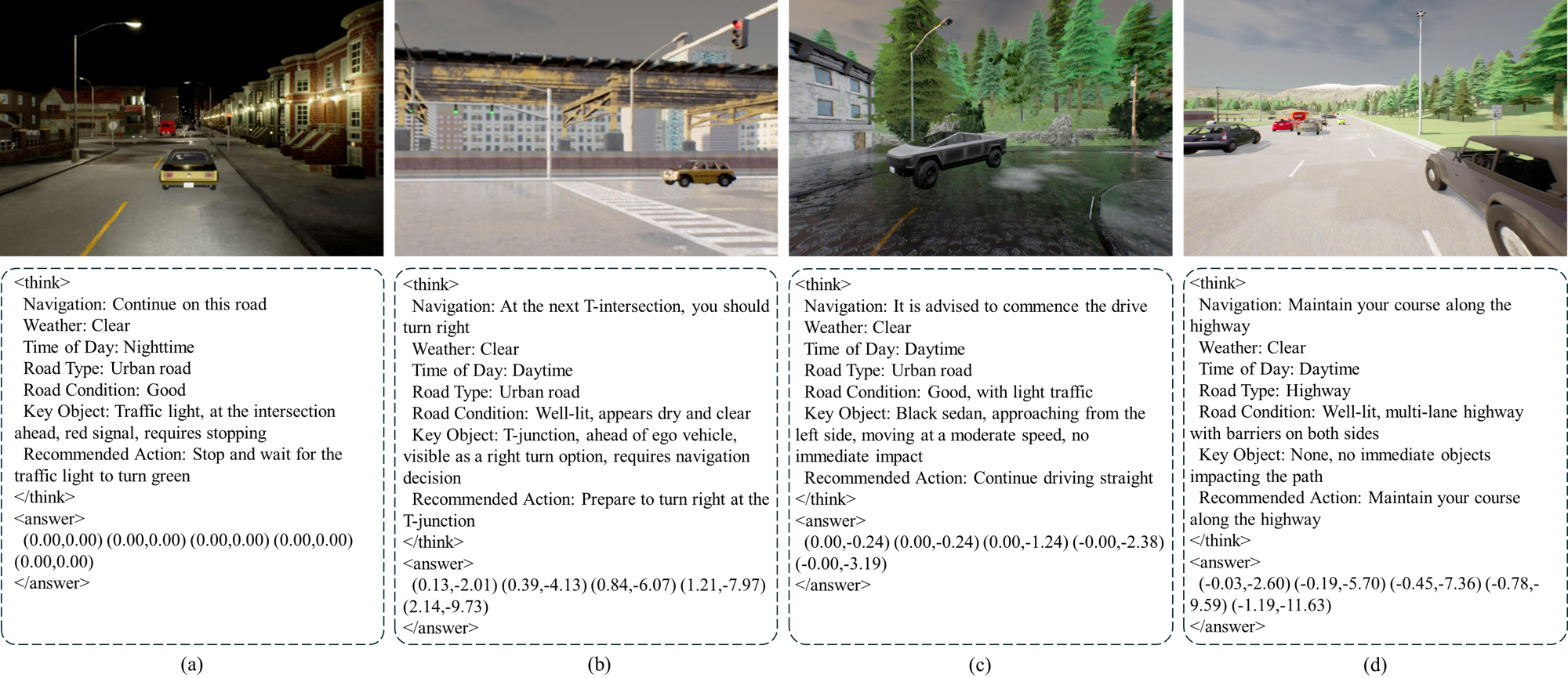}
    \caption{The explicit think-and-answer reasoning process for representative driving scenarios from the CARLA simulator. This includes urban road and highway settings, as well as daytime and nighttime conditions.}
    \label{fig:cl_scenarios}
\end{figure*}

    \subsection{Qualitative Analysis}
        \subsubsection{Think-and-Answer Reasoning Process}
    We demonstrate the reasoning process with representative scenarios from CARLA in Fig.~\ref{fig:cl_scenarios}. The top panel displays selected keyframes for each scenario, providing a visual context of the driving environment. The bottom panel includes detailed reasoning content. 
    Scenario (a) depicts a nighttime urban setting with clear weather conditions. In this low-light condition, DSDrive manages to identify the traffic signal and execute the required action of stopping. 
    Scenario (b) requires the vehicle to make a right turn at a T-intersection, where the vehicle needs to accurately identify the correct traffic signal among multiple visible signals, each displaying different indications for its turning options. DSDrive effectively discerns the appropriate traffic light to execute the right turn, demonstrating its adeptness in parsing complex visual information and adhering to traffic rules.
    Scenario (c) features a suburban environment with incoming vehicles on the closely adjacent lane. DSDrive effectively identifies the black sedan approaching from the left side.
    Scenario (d) takes place in a highway environment with relatively heavy traffic. Despite the congestion, DSDrive effectively identifies that none of the surrounding vehicles are directly impacting its path, allowing it to maintain a steady and stable course.
    In the aforementioned scenarios, DSDrive demonstrates robust performance in recognizing traffic signals, executing turns, and navigating a wide range of environments, from urban streets to highway settings. The explicit think-and-answer reasoning process shows that DSDrive manages to learn advanced reasoning capability from the large-sized VLM and applies it within the E2E AD framework. This consistent performance across diverse scenarios underscores the effectiveness and reliability of DSDrive in interpreting visual cues and making informed decisions. More importantly, through the waypoint-driven dual-head coordination module, the LLM within DSDrive exhibits a critical ability to accurately predict waypoints that adapts well to dynamic traffic situations. This capability is essential for delivering the final planning phase of AD systems.
    
        \begin{figure}[t]
            \centering
            \includegraphics[trim=0 0 0 0, clip, width=1\linewidth]{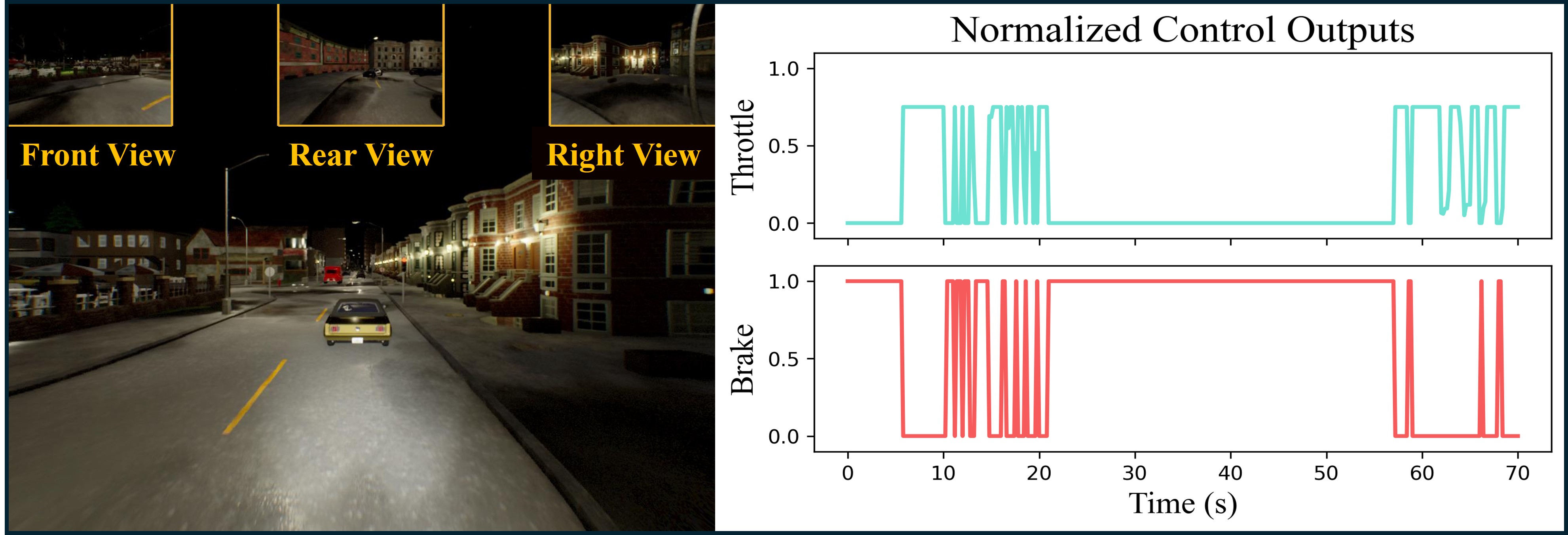}
            \caption{Performance at a typical urban intersection in response to traffic light signals. The AV first stops at a red light and subsequently starts driving when the light turns green (left). The visualizations of the throttle and brake control inputs of the AV (right) demonstrate its prompt deceleration and acceleration in accordance with the traffic signal phase.}
            \label{fig:light}
        \end{figure}

        \begin{figure}[t]
            \centering
            \includegraphics[trim=0 0 0 0, clip, width=1\linewidth]{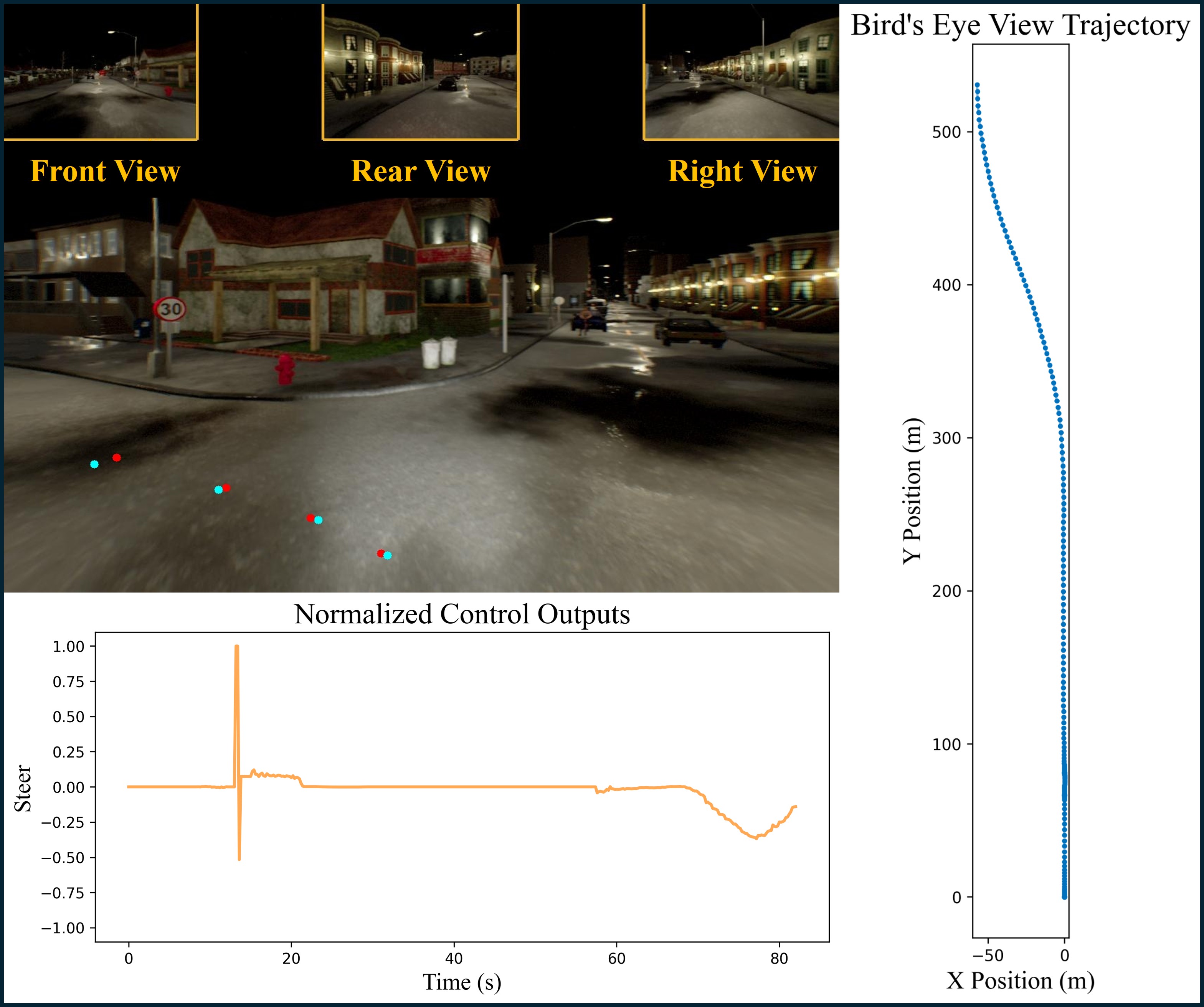}
            \caption{Performance when the AV approaches an intersection and makes a left turn. The image (top left) shows the predicted waypoints from the prediction heads in planning (red dots) and reasoning (cyan dots). The visualizations of the steering (bottom left) and BEV trajectory (right) are included to validate the efficacy of closed-loop planning and control.}
            \label{fig:left_turn}
        \end{figure}

        \begin{figure}[t]
            \centering
            \includegraphics[trim=0 0 0 0, clip, width=1\linewidth]{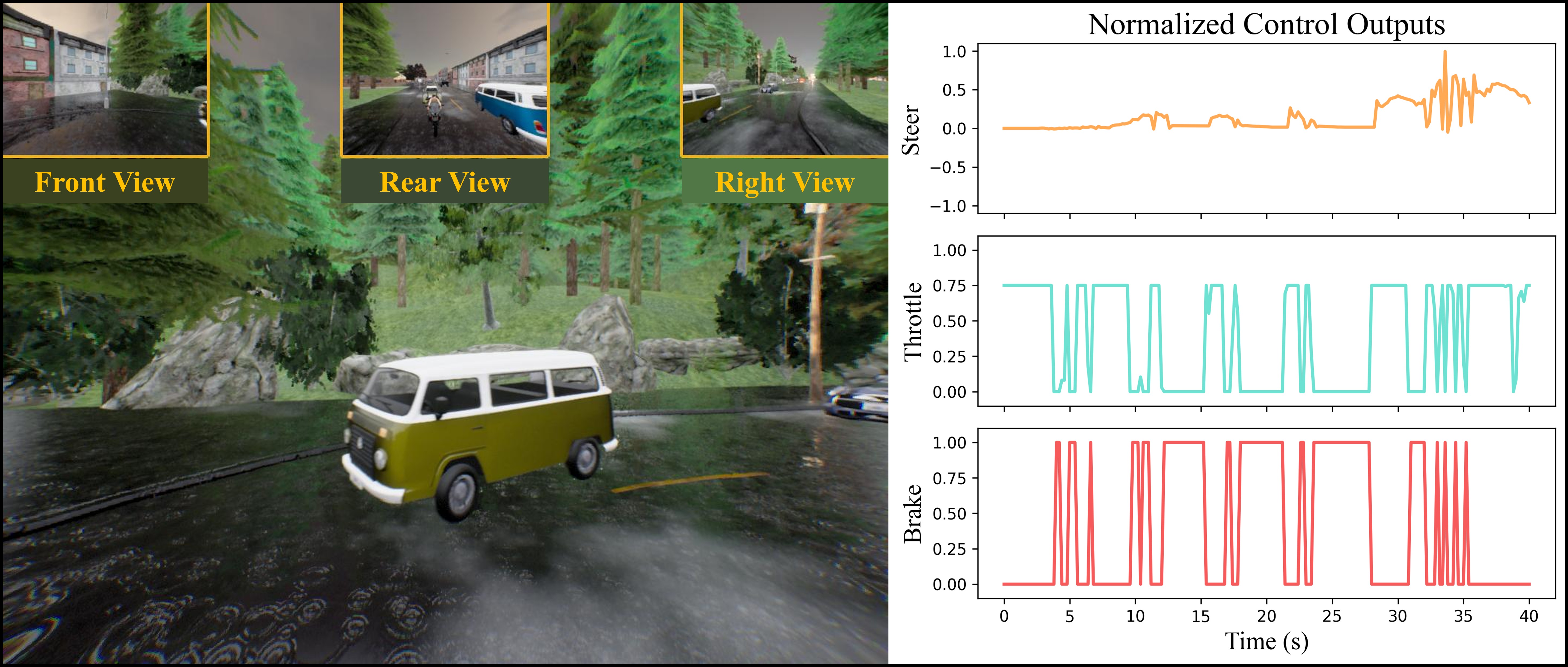}
            \caption{
            Performance of the AV in consecutively yielding to approaching vehicles on a curved roadway (left). The visualizations of the steering, throttle, and brake control inputs (right) demonstrate the AV's capability to identify objects with potential collision risks and react with informed planning and control strategies.}
            \label{fig:yielding}
        \end{figure}

        \begin{figure}[t]
            \centering
            \includegraphics[trim=0 0 0 0, clip, width=1\linewidth]{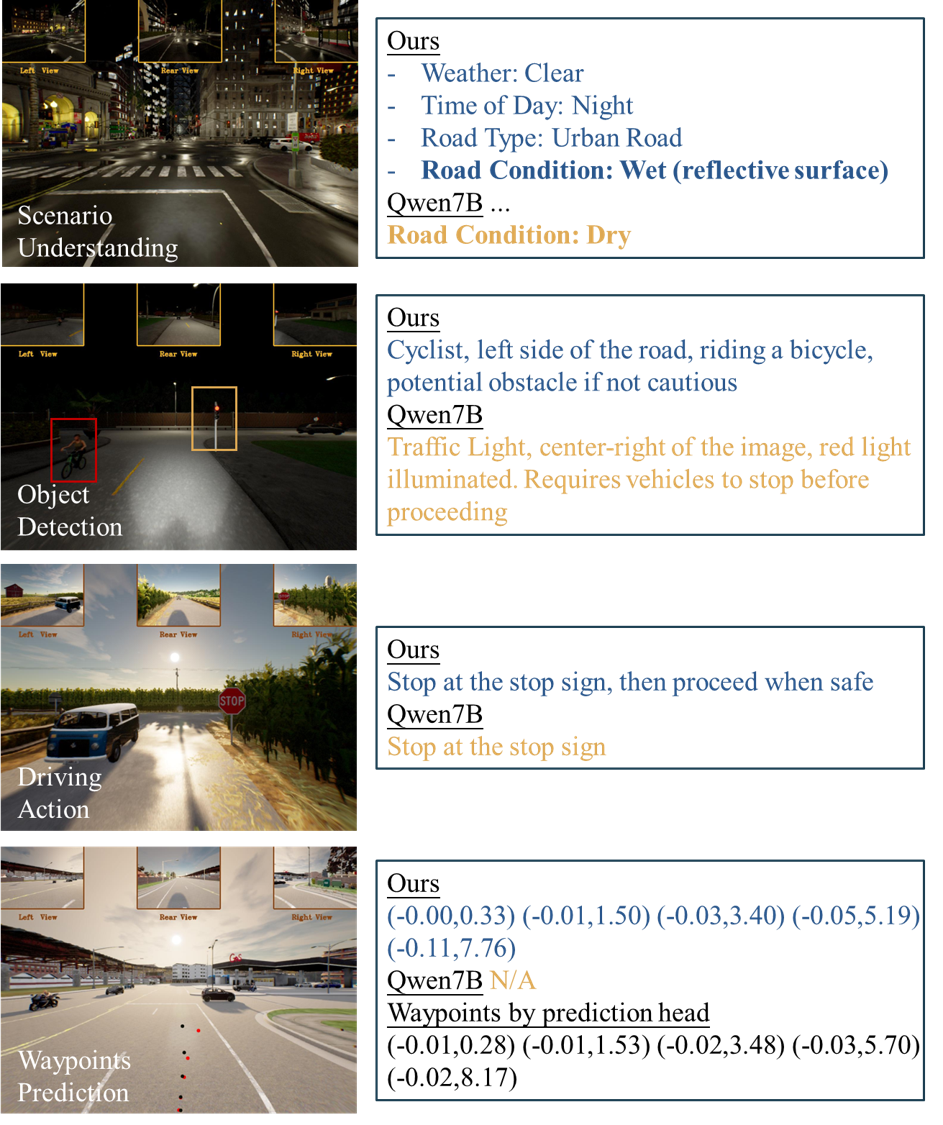}
            \caption{Comparison of reasoning capabilities of our method to Qwen2.5-VL-7B. Our method demonstrates superior accuracy and context enrichment in answers, including enhanced waypoint prediction abilities.}
            \label{fig:ol_answer}
        \end{figure}



\begin{figure*}[htbp]
\centering
\begin{minipage}{0.24\linewidth}
\vspace{1pt}
\centerline{\includegraphics[width=\textwidth]{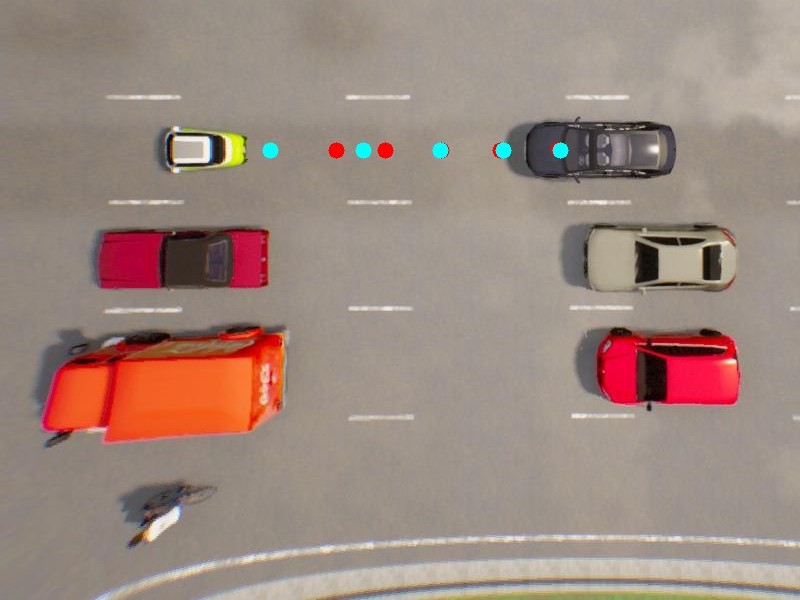}}
\centerline{\footnotesize$t=0.85\,\mathrm{s}, v=2.29\, \mathrm{m/s}$}
\end{minipage}
\begin{minipage}{0.24\linewidth}
\vspace{1pt}
\centerline{\includegraphics[width=\textwidth]{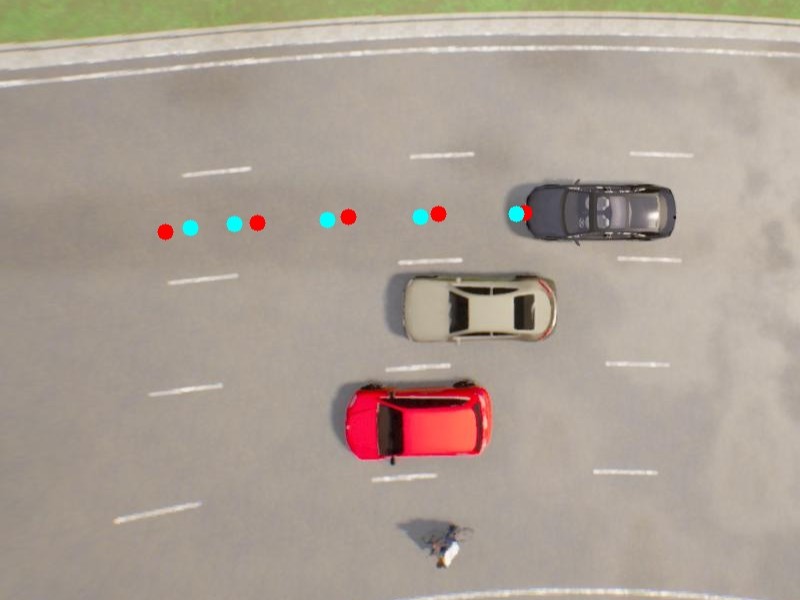}}
\centerline{\footnotesize$t=5.00\,\mathrm{s}, v=4.41\, \mathrm{m/s}$}
\end{minipage}
\begin{minipage}{0.24\linewidth}
\vspace{1pt}
\centerline{\includegraphics[width=\textwidth]{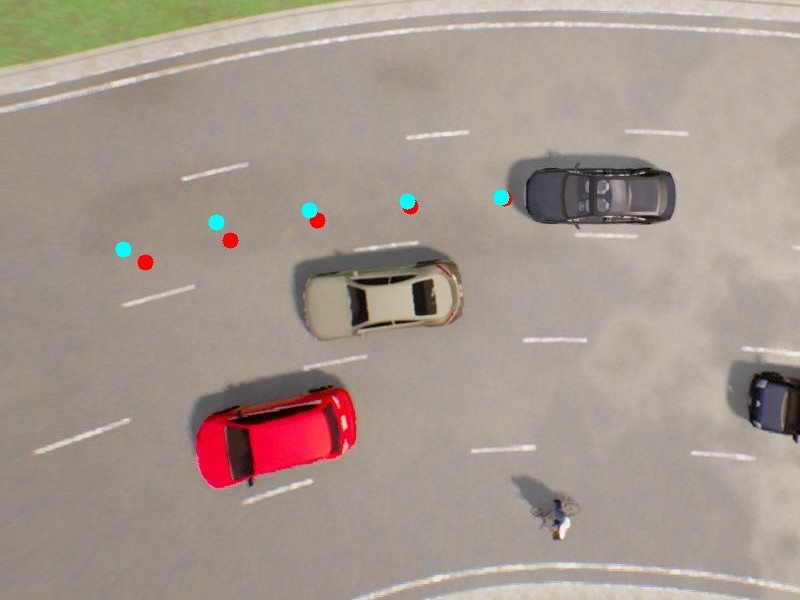}}
\centerline{\footnotesize$t=7.95\,\mathrm{s}, v=6.16\, \mathrm{m/s}$}
\end{minipage}
\begin{minipage}{0.24\linewidth}
\vspace{1pt}
\centerline{\includegraphics[width=\textwidth]{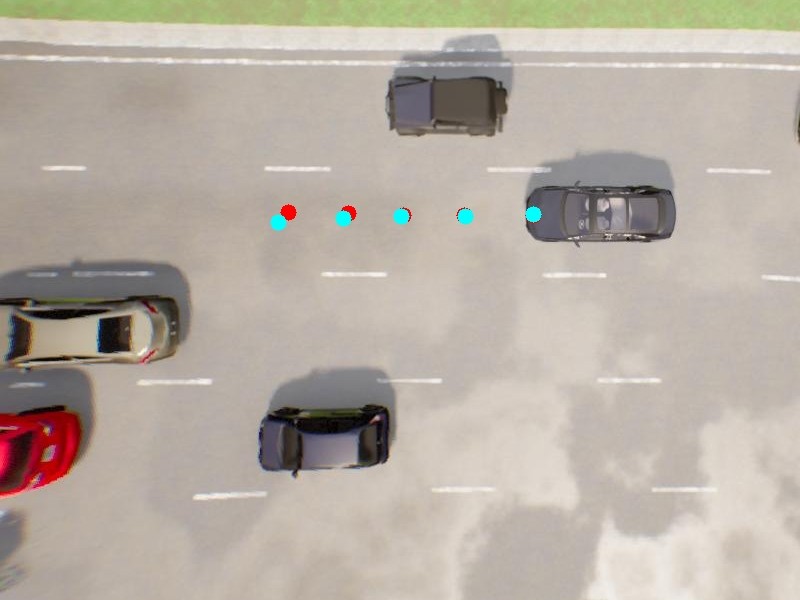}}
\centerline{\footnotesize$t=13.4\,\mathrm{s}, v=3.98\, \mathrm{m/s}$}
\end{minipage}
\begin{minipage}{0.9\linewidth}
\vspace{6pt}
\centerline{\footnotesize (a) Highway with moderately heavy traffic.}
\vspace{3pt}
\end{minipage}

\begin{minipage}{0.24\linewidth}
\vspace{1pt}
\centerline{\includegraphics[width=\textwidth]{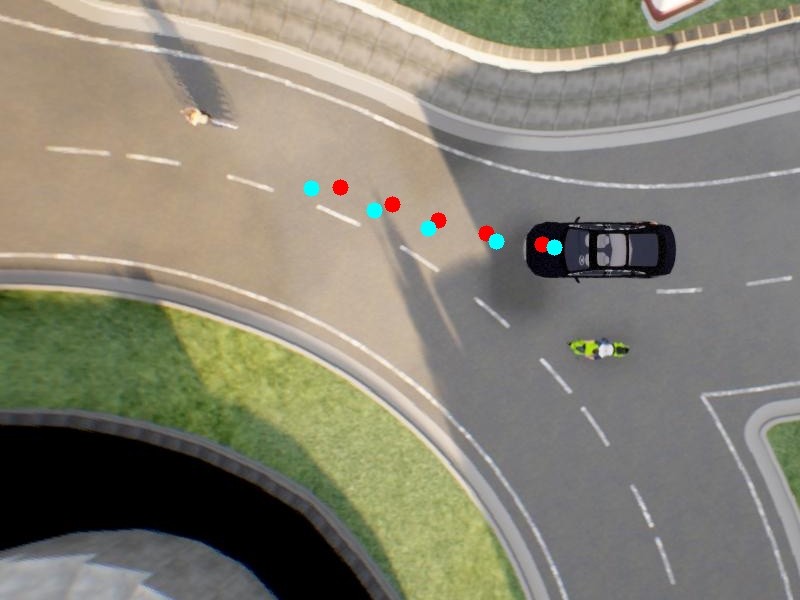}}
\centerline{\footnotesize$t=11.45\,\mathrm{s}, v=4.03\, \mathrm{m/s}$}
\end{minipage}
\begin{minipage}{0.24\linewidth}
\vspace{1pt}
\centerline{\includegraphics[width=\textwidth]{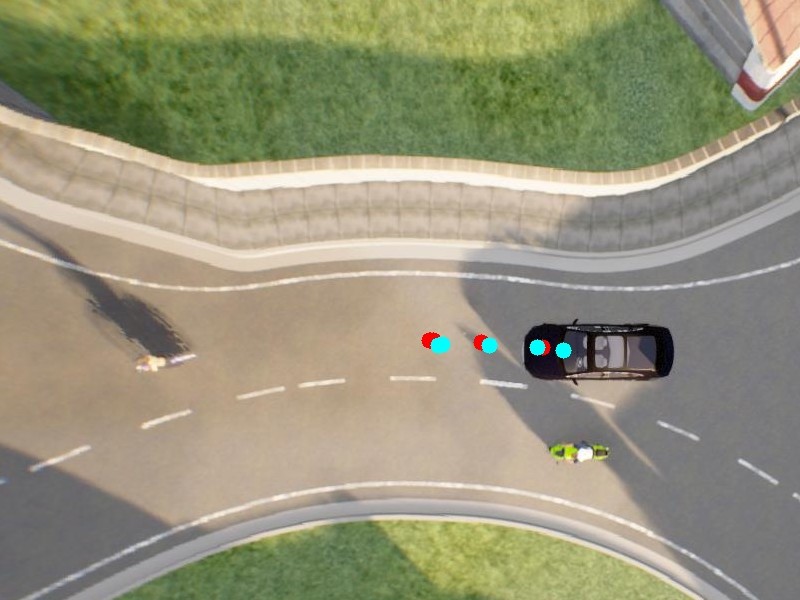}}
\centerline{\footnotesize$t=12.2\,\mathrm{s}, v=3.37\, \mathrm{m/s}$}
\end{minipage}
\begin{minipage}{0.24\linewidth}
\vspace{1pt}
\centerline{\includegraphics[width=\textwidth]{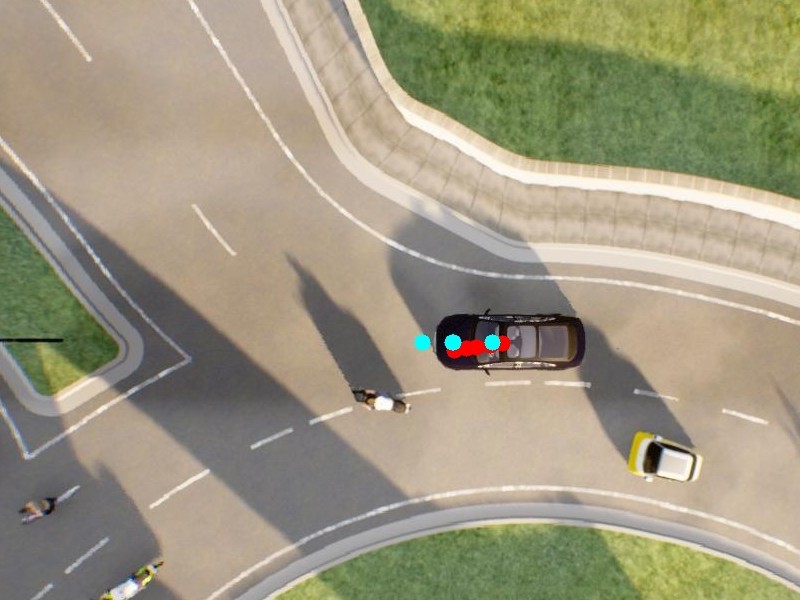}}
\centerline{\footnotesize$t=15.25\,\mathrm{s}, v=0.56\, \mathrm{m/s}$}
\end{minipage}
\begin{minipage}{0.24\linewidth}
\vspace{1pt}
\centerline{\includegraphics[width=\textwidth]{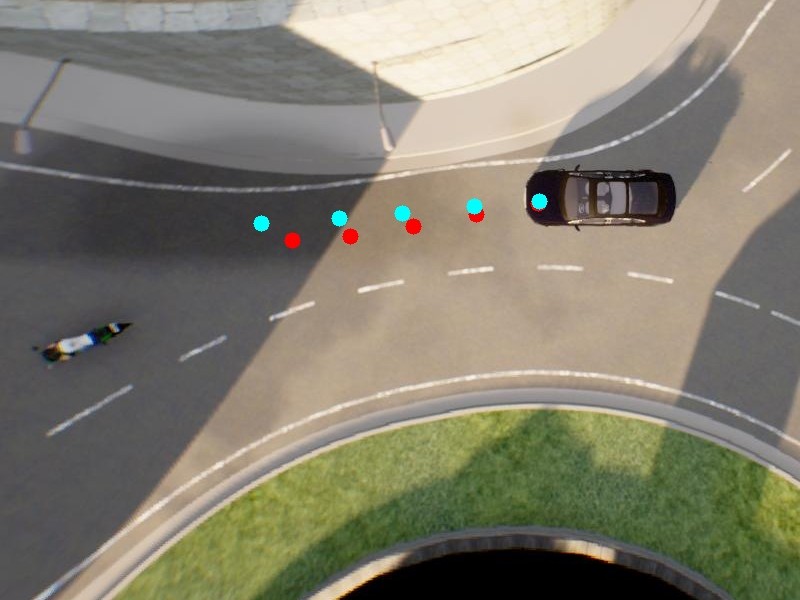}}
\centerline{\footnotesize$t=17.20\,\mathrm{s}, v=4.21\, \mathrm{m/s}$}
\end{minipage}
\begin{minipage}{0.9\linewidth}
\vspace{6pt}
\centerline{\footnotesize (b) Roundabout.}
\vspace{3pt}
\end{minipage}

\begin{minipage}{0.24\linewidth}
\vspace{1pt}
\centerline{\includegraphics[width=\textwidth]{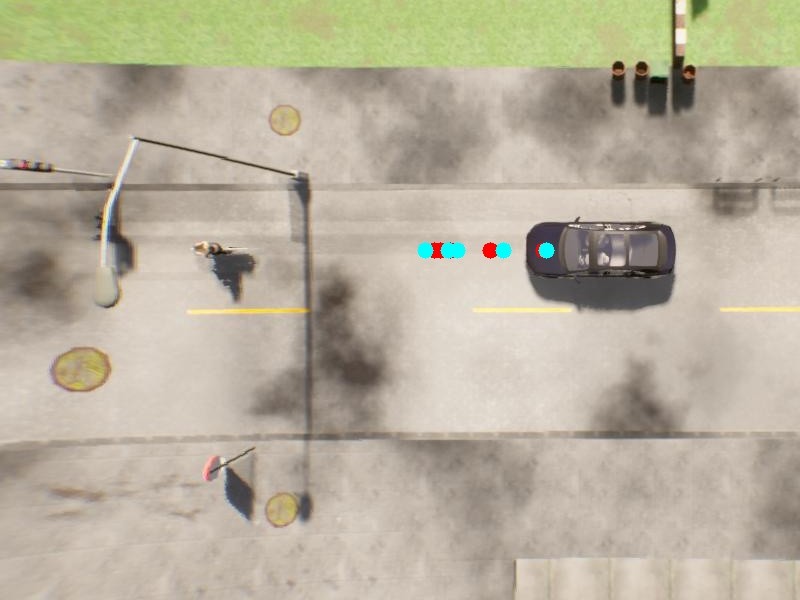}}
\centerline{\footnotesize$t=3.65\,\mathrm{s}, v=3.79 \, \mathrm{m/s}$}
\end{minipage}
\begin{minipage}{0.24\linewidth}
\vspace{1pt}
\centerline{\includegraphics[width=\textwidth]{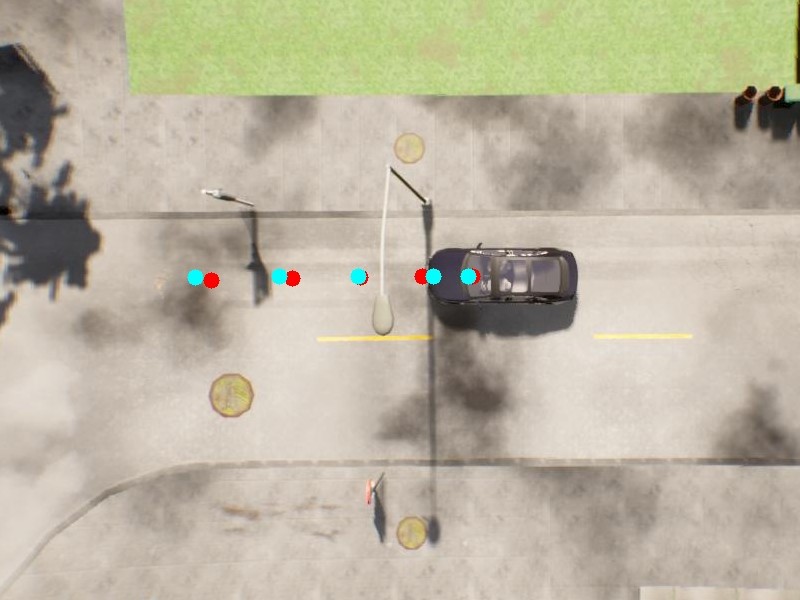}}
\centerline{\footnotesize$t=14.45\,\mathrm{s}, v=0.97\, \mathrm{m/s}$}
\end{minipage}
\begin{minipage}{0.24\linewidth}
\vspace{1pt}
\centerline{\includegraphics[width=\textwidth]{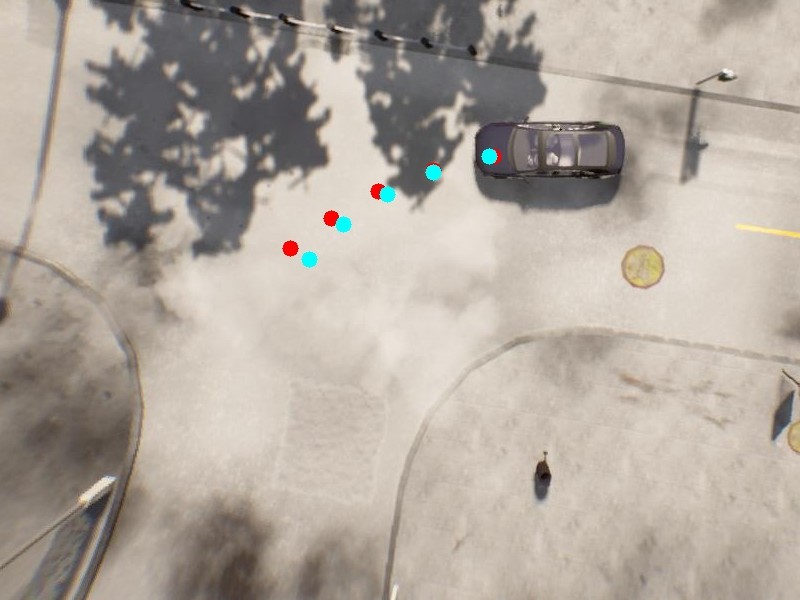}}
\centerline{\footnotesize$t=16.05\,\mathrm{s}, v=2.59\, \mathrm{m/s}$}
\end{minipage}
\begin{minipage}{0.24\linewidth}
\vspace{1pt}
\centerline{\includegraphics[width=\textwidth]{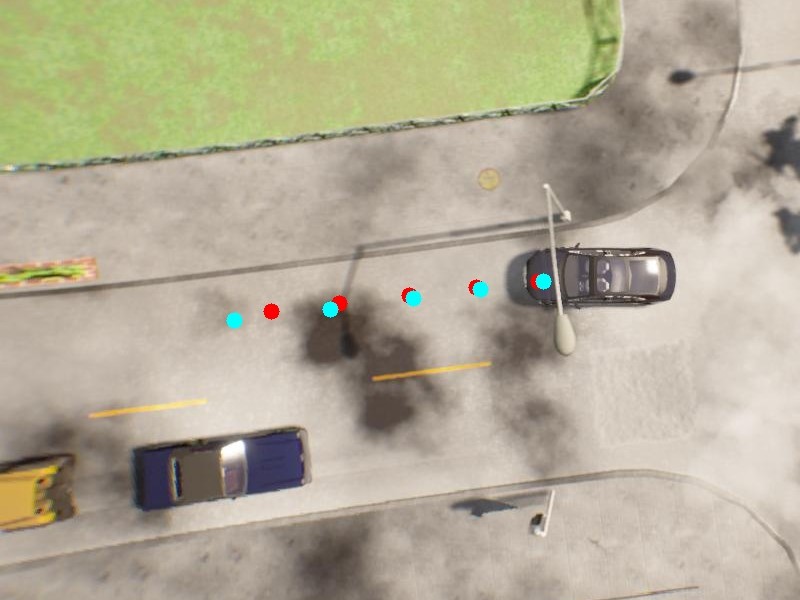}}
\centerline{\footnotesize$t=18.80\,\mathrm{s}, v=4.13\, \mathrm{m/s}$}
\end{minipage}
\begin{minipage}{1.0\linewidth}
\vspace{6pt}
\centerline{\footnotesize(c) Urban driving with an intersection.}
\end{minipage}
\caption{Visualization of the reasoning and planning alignment in CARLA. As illustrated across three distinct operational scenarios: (a) Highway with moderately heavy traffic; (b) Roundabout; (c) Urban driving with an intersection. The black vehicle represents the AV. The waypoints generated by the reasoning head are shown in cyan, and those predicted by the planning head are shown in red. $t$ denotes time and $v$ represents velocity.}
\label{fig:waypt}
\end{figure*}

    \subsubsection{Representative Scenarios}
    Representative scenarios are included to provide insights to the ability of DSDrive in closed-loop driving, as this is the goal of this study.
    The ability to manage the traffic light is discussed in Fig.~\ref{fig:light}. Firstly, DSDrive is able to detect the red light and stop at a safe distance from the vehicle ahead. Once the traffic light changes to green, DSDrive seamlessly transitions from a stationary position to active driving. Subsequently, as indicated by Fig.~\ref{fig:left_turn}, the vehicle executes a left turn as directed by the navigation instruction. This sequence underscores DSDrive's capability to accurately interpret and execute navigation commands, reflecting its effective integration of reasoning and operational execution in complex urban environments. 
    In the suburban area with approaching vehicles nearby in Fig.~\ref{fig:yielding}, DSDrive yields appropriately to the incoming traffic, ensuring safety while smoothly navigating through the curved road. 
    Visualizations of the key control variables pertinent to the planning process are included for the aforementioned scenarios, which demonstrate DSDrive's ability to integrate situational awareness with precise control.

    \subsubsection{Reasoning Capability}
    We further assess the quality of answers to provide insights into the reasoning capability of DSDrive. We compare the answers generated by DSDrive and Qwen2.5-VL-7B across the dimensions of scenario understanding, object detection, driving actions, and waypoints prediction in Fig.~\ref{fig:ol_answer}. Despite its smaller model size (1B) as compared to Qwen2.5-VL-7B, DSDrive does not fall short in the designated reasoning abilities for AD. In certain aspects, it outperforms Qwen2.5-VL-7B. For example, in scenario understanding, DSDrive accurately recognizes the wet road condition with a reflective surface, whereas Qwen2.5-VL-7B fails to identify this detail. In object detection, DSDrive attends to the cyclist on the left side and considers its potential impact on the vehicle's path. In driving actions, DSDrive considers the further maneuver to proceed when safe, in addition to stopping at the stop sign, while Qwen2.5-VL-7B terminates the answer at stopping at the stop sign. Notably, DSDrive is capable of predicting waypoints through a text-based reasoning process, aligning with the purpose of the waypoints predictor in the \EE\ AD framework, while Qwen2.5-VL-7B is not explicitly tuned to support this functionality. This difference underscores the effectiveness of our alignment between the semantic reasoning task and the numerical trajectory planning task, achieved through the purposeful design of both the training dataset and the model architecture.

\subsection{Dual-Head Coordination}
    The innovations of the waypoint-driven dual-head coordination module in DSDrive are twofold. Firstly, the waypoints are embedded in the training dataset for the distillation of reasoning capability within the proposed framework, creating a shared goal for the two tasks. Secondly, the \EE\ driving model is built with interlinked prediction heads for reasoning and planning. The two closely related tasks thereby complement each other in that the reasoning outputs provide explanation for the planning outcome, and the planning head can generate trajectory prediction akin to the reasoning head, but with faster inference for efficiency in deployment.

    This investigation raises the following questions: (1) Does the dual-head coordination module acts effectively for the alignment of reasoning and planning tasks as intended? (2) Does this design enhance the performance of DSDrive as an AD agent in closed-loop driving scenarios? 
    
    Focusing on the first question, Fig.~\ref{fig:waypt} provides a qualitative demonstration through a direct visualization of the alignment between waypoints predicted by the reasoning and planning heads. 
    Scenario (a) depicts a highway with moderately heavy traffic. The predicted waypoints generated by both the reasoning and planning heads display similar movement trajectories. This enables the AV to follow the vehicle ahead, navigate through curved roads, and maintain smooth cruising within the traffic flow.
    Scenario (b) depicts a roundabout scenario where the AV exhibits the following behaviors, as evidenced by the predicted waypoints: entering the roundabout, maintaining normal driving within the roundabout, and yielding to a cyclist crossing the path. In this yielding instance, the waypoints predicted by both the reasoning and planning heads suggest a stopping tendency to ensure safe interaction with the cyclist. Subsequently, the AV accelerates back to normal speed once the path is clear of obstacles.
    Scenario (c) depicts a typical urban driving scenario with an intersection. Initially, the predicted waypoints reflect a deceleration phase as the AV approaches a red traffic light. Subsequently, as the traffic light transitions to green, the waypoints indicate an acceleration phase, enabling the AV to proceed through the intersection. Following this, the predicted waypoints generated by both the reasoning and planning heads correspond to the task of executing a left turn at the intersection. Finally, these waypoints align with the subsequent cruising phase.
    

    Regarding the second question, we further provide an ablation study for evaluating the impact of the dual-head coordination module on driving performance. We implement closed-loop simulations in the LangAuto benchmarks across tiny, short, and long routes, and the experiment results are summarized in Table~\ref{tab:dual}. Dual-task refers to our method. Two ablations on the training dataset are considered: CoT only uses the answer generated from the large-sized VLM, without embedding of ground-truth data such as navigation commands or waypoints. GT only, on the other hand, is built with only ground-truth data, but with no inputs from the large-sized VLM. It appears that our method beats the two ablatives with higher DS, RC, and IS, supporting the superiority of the waypoint-driven dual-head coordination module design in facilitating \EE\ closed-loop AD.

\begin{table}[t]
  \centering
  \caption{Ablations on the waypoint-driven dual-head coordination module design.}
    \begin{tabular}{ccccc}
    \toprule
    Benchmark & Train Method & {DS $\uparrow$} & {RC $\uparrow$} & {IS $\uparrow$} \\
    \midrule
    \multirow{3}[2]{*}{LangAuto-Long} & CoT only & 23.56  & 27.08  & 0.88  \\
          & GT only & 21.91  & 26.51  & 0.85  \\
          & Dual-task & \textbf{27.35}  & \textbf{29.52}  & \textbf{0.89}  \\
    \midrule
    \multirow{3}[2]{*}{LangAuto-Short} & CoT only & 38.42  & 44.99  & 0.83  \\
          & GT only & 40.90  & 49.72  & \textbf{0.86}  \\
          & Dual-task & \textbf{42.91}  & \textbf{51.76}  & 0.85  \\
    \midrule
    \multirow{3}[2]{*}{LangAuto-Tiny} & CoT only & 46.14  & 54.46  & 0.87  \\
          & GT only & 48.50  & \textbf{55.83}  & 0.88  \\
          & Dual-task & \textbf{49.35}  & 54.50  & \textbf{0.89}  \\
    \bottomrule
    \end{tabular}%
  \label{tab:dual}%
\end{table}%

\subsection{Computation Efficiency}

    Designed as a compact LLM specifically tailored for \EE\ AD systems, the primary objective of DSDrive is to optimize computational efficiency by reducing both inference time and memory consumption, thereby enhancing the feasibility of deploying such systems in real-world applications with limited computational resources.
    As shown in Table~\ref{tab:time}, we conduct a comparative analysis against existing models, specifically LMDrive implemented with LLaVA-7B and LLaMA-1B, to understand the inference resource requirement of DSDrive.
    
    The results highlight the comparative computational efficiency of DSDrive. In terms of inference time, the performance of DSDrive is at a comparable level with the LMDrive model utilizing LLaMA-1B and is slightly faster than the LLaVA-7B. In terms of memory usage, DSDrive exhibits a peak memory requirement of 8082 MB. While this represents an increase compared to the LMDrive with LLaMA-1B (6682 MB), it remains significantly lower than the LLaVA-7B model's consumption of 14263 MB. The modest increase in memory usage relative to LLaMA-1B is justified by the enhanced capability and robustness that DSDrive offers, suggesting a balanced trade-off between memory efficiency and model performance.
    
    The implications of these results are twofold. Firstly, the comparable inference time between DSDrive and the LMDrive series suggests that the integration of reasoning capability within DSDrive's architecture does not significantly compromise the processing speed of the \EE\ AD system. Secondly, the memory efficiency of DSDrive, especially when contrasted with LLaVA-7B, underscores its suitability for deployment in environments where hardware constraints are a critical consideration. It is important to note that the modest memory requirements of DSDrive make it feasible for deployment on advanced embedded infrastructures such as the NVIDIA Orin platform, facilitating broader adoption of AD technology.

    \begin{table}[t]
      \centering
      \caption{Computational time and memory required for inference.}
        \begin{tabular}{ccc}
        \toprule
              & Inference time (s) & Peak memory (MB) \\
        \midrule
        LMDrive (LLaVA-7B) & 0.06  & 14263 \\
        LMDrive (LLaMA-1B) & 0.05  & 6682 \\
        DSDrive (LLaMA-1B) & 0.05  & 8082 \\
        \bottomrule
        \end{tabular}%
      \label{tab:time}%
    \end{table}%


\section{Conclusion} \label{sec:con}
In this paper, we present DSDrive, an E2E AD framework leveraging a compact LLM. This framework features a knowledge distillation process to facilitate the deployment of small-sized LLMs, which lead to efficient computation while enhancing their capabilities. Furthermore, it manages to bridge the gap between high-level reasoning and low-level planning through a waypoint-driven dual-head coordination module. 
We conduct extensive closed-loop simulation experiments to evaluate the performance of DSDrive. The results demonstrate that DSDrive achieves driving performance comparable to that of larger systems and even outperforms them in several key metrics, despite its compact nature. Also, DSDrive successfully navigates representative scenarios, including traffic lights, intersections, and heavy traffic conditions. 
The efficacy of the dual-head coordination module is also thoroughly investigated. The findings indicate that this module effectively aligns reasoning and planning tasks by utilizing waypoints as a shared objective, thereby enhancing overall driving performance. 
Additionally, DSDrive reduces inference time and memory usage during deployment, offering a practical solution for integration into AD systems with constrained computational resources.

Future research could evolve along two interesting directions. Firstly, we can focus on enhancing the capability of compact-sized LLMs using methods such as reinforcement learning or contrastive learning. Secondly, efforts should be directed toward further reducing the size of compact-sized LLMs to enable the deployment of LLM-based E2E AD systems in real vehicles as well as other edge infrastructures.

\bibliographystyle{IEEEtran}
\bibliography{mybibfile}

\end{document}